\documentclass{egpubl}
\usepackage{eg2025}

\ConferencePaper

\CGFccby

\usepackage[T1]{fontenc}
\usepackage{dfadobe} 

\electronicVersion

\usepackage{cite}
\usepackage{hyperref}
\usepackage{egweblnk} 

\usepackage{microtype}
\usepackage{graphicx}
\usepackage{booktabs} %

\usepackage{xcolor}
\usepackage{xspace}
\usepackage{bm}
\usepackage[normalem]{ulem} %

\usepackage{algorithm} %
\usepackage{algpseudocode}
\usepackage{amsfonts}
\usepackage{dsfont}
\usepackage{mathtools}
\usepackage{tabularx}
\usepackage{multirow}

\usepackage{makecell}

\usepackage[capitalize]{cleveref}
\crefname{section}{Sec.}{Secs.}
\Crefname{section}{Section}{Sections}
\Crefname{table}{Table}{Tables}
\crefname{table}{Tab.}{Tabs.}

\makeatletter
\DeclareRobustCommand\onedot{\futurelet\@let@token\@onedot}
\def\@onedot{\ifx\@let@token.\else.\null\fi\xspace}

\def\ie{\emph{i.e}\onedot}

\def\etal{\emph{et al}\onedot}
\makeatother

\let\citet\cite

\newcommand{\EFGVI}{E\textsuperscript{2}FGVI}

\title[Infusion]{Infusion: Internal Diffusion for Inpainting of Dynamic Textures and Complex Motion}

\author[N. Cherel, A. Almansa, Y. Gousseau, A. Newson]
{\parbox{\textwidth}{\centering N. Cherel$^{1}$, A. Almansa$^2$, Y. Gousseau$^1$, A. Newson$^3$}
\\
{\parbox{\textwidth}{\centering $^1$LTCI, Télécom Paris, Institut Polytechnique de Paris, France\\
         $^2$MAP5, CNRS \& Université Paris Cité, France \\
         $^3$ISIR, Sorbonne Université, France 
         }
}
}

\begin{document}

\teaser{
    \centering
    \includegraphics[width=\textwidth]{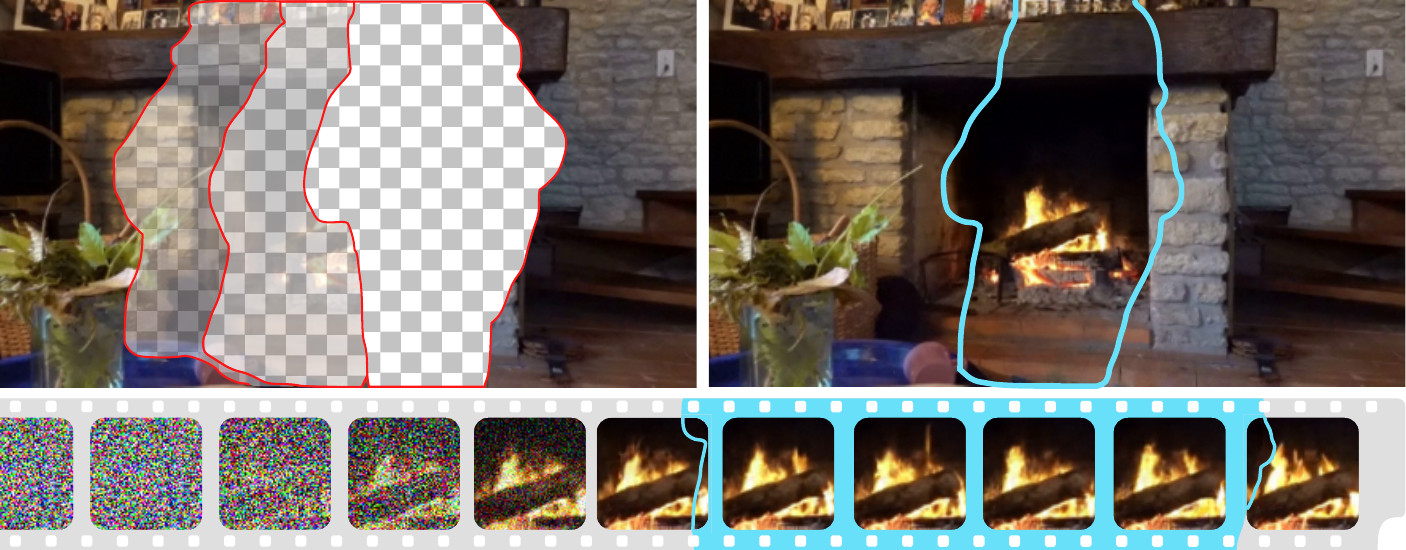}
    \caption{Our method can inpaint videos with dynamic textures and complex motion using a diffusion-based approach (inpainted region in blue in the film reel). High quality results are obtained with a reasonable network size by learning on the input video only, and leveraging our proposed training strategy: interval training. Training and inference are mixed during the diffusion process, which is divided into intervals.}
    \label{fig:teaser}
}

\maketitle

\begin{abstract}
Video inpainting is the task of filling a region in a video in a visually convincing manner. It is very challenging due to the high dimensionality of the data and the temporal consistency required for obtaining convincing results. Recently, diffusion models have shown impressive results in modeling complex data distributions, including images and videos.
Such models remain nonetheless very expensive to train and to perform inference with, which strongly
reduce their applicability to videos, and yields unreasonable computational loads. We show that in the case of video inpainting, thanks to the highly auto-similar nature of videos, the training data of a diffusion model can be restricted to the input video and still produce very satisfying results. With this internal learning approach, where the training data is limited to a single video, our lightweight models perform very well with only half a million parameters, in contrast to the very large networks with billions of parameters typically found in the literature. We also introduce a new method for efficient training and inference of diffusion models in the context of internal learning, by splitting the diffusion process into different learning intervals corresponding to different noise levels of the diffusion process. 
We show qualitative and quantitative results, demonstrating that our method reaches or exceeds state of the art performance in the case of dynamic textures and complex dynamic backgrounds.


\end{abstract}

\section{Introduction}

Video inpainting, or video completion, aims to fill in an unknown region in a video, in a visually convincing manner. While in the case of images, very impressive results have been obtained using deep learning approaches~\cite{yuGenerativeImageInpainting2018,sahariaPaletteImagetoImageDiffusion2022b}, the extension of these methods to video has proven to be extremely difficult. This is due to the great technical challenges of videos, such as the large dimensionality of the data which leads to computationally expensive methods, the high sensitivity of the human eye to temporal inconsistencies, and the difficulty of inpainting complex motions. Many current state of the art methods combine deep learning with optical flow to deal with temporal consistency and motion~\cite{liEndtoEndFrameworkFlowGuided2022a,zhouProPainterImprovingPropagation2023a}. Unfortunately, optical flow breaks down in the presence of complex motions, for example in the case of dynamic (video) textures such as waves or tree leaves, which are poorly synthesized by such methods, as we show in this paper.

Concurrently, diffusion models have recently emerged as a powerful framework to produce samples of complex distributions, allowing the synthesis of realistic and diverse examples of images, with unprecedented quality~\cite{hoDenoisingDiffusionProbabilistic2020b}. However, these networks are extremely large (up to 500 million parameters for image inpainting \cite{Lugmayr_2022_CVPR,sahariaPaletteImagetoImageDiffusion2022b}), leading to very long training and inference times. While several acceleration techniques have been proposed~\cite{songDenoisingDiffusionImplicit2021,rombachHighResolutionImageSynthesis2022b} in the case of images, they would nevertheless yield prohibitive training times for video inpainting. Furthermore, while the common practice of fine-tuning large diffusion models for specific tasks is feasible in the case of images, it is still not practical in the case of videos due to the large size of video diffusion models.

In this paper, we propose a diffusion-based video inpainting method which has both a reasonable network size (around 500k parameters), and avoids relying on optical flow that can lead to poor reconstruction of complex motions. We show that a lightweight diffusion network, combined with an ``internal learning'' method, produces high-quality video inpainting results, especially in the difficult case of complex motion. Internal learning refers to restricting the training database to the input video, which makes sense, due to the highly auto-similar nature of videos~\cite{wexlerSpaceTimeCompletionVideo2007}. In particular, we propose a novel training algorithm for diffusion models, referred to as ``interval training'', which significantly improves the performance in this context. This allows us to generate high quality results with a lightweight network, leading to feasible training and inference times (between 3-15 hours on a single GPU). 
In contrast to most of the other video inpainting algorithms which mainly rely on optical flow, our method handles dynamic textures well and is able to produce diverse inpainting solutions. Finally, temporal coherence is ensured by 3D convolutions in the network.

We evaluate our method quantitatively and qualitatively on several databases, displaying its ability to inpaint complex motions such as dynamic textures and moving objects. We show that optical flow-based methods fail in the case of dynamic textures, due to the stochastic nature of such motion. Our method, on the other hand, is purely based on diffusion, which is designed precisely to sample complex distributions. This confirms that diffusion is a good choice for such stochastic motion components.

\section{Related work}
\label{sec:relatedWork}

\subsection{Diffusion Models}
Diffusion models~\cite{sohl-dicksteinDeepUnsupervisedLearning2015a,hoDenoisingDiffusionProbabilistic2020b} have shown great generative capabilities on image datasets.
They have supplanted Generative Adversarial Networks (GAN) in quality and training stability~\cite{dhariwalDiffusionModelsBeat2021}.
One major drawback, however, is the slow inference and training, which has lead to several attempts at acceleration up with various techniques~\cite{rombachHighResolutionImageSynthesis2022b,nicholImprovedDenoisingDiffusion2021a}.

Balaji~\etal~\citet{balajiEDiffITexttoImageDiffusion2023a} propose to create an ensemble of diffusion networks specialised in different steps of the diffusion process. This approach shares some similarities with our training method (it was developed concurrently to our work), however a major difference is that theirs does not address the issue of network size, which is so crucial in our video inpainting problem.

For image inpainting, Lugmayr~\etal~\citet{Lugmayr_2022_CVPR} use an unconditional diffusion model, with projections and time-travels for better coherence. Saharia~\etal~\citet{sahariaPaletteImagetoImageDiffusion2022b} train a conditional model for diverse image-to-image problems.
User-guided inpainting is possible with text~\cite{xieSmartBrushTextShape2022} or other modalities (strokes, examples)~\cite{yangUnipaintUnifiedFramework2023}.

However, the extension of diffusion models to videos is not straightforward, in particular due to their huge sizes and training times.
Blattmann~\etal~\citet{blattmannAlignYourLatents2023a} finetune an image model for videos by introducing temporal attention layers.
Another option is to decompose the video as multiple images via neural atlases~\cite{kastenLayeredNeuralAtlases2021a} which can then be handled by image-based diffusion models~\cite{chaiStableVideoTextdrivenConsistencyaware2023a}.
Ceylan~\etal~\citet{ceylanPix2VideoVideoEditing2023} use noise inversion and attention maps for consistent video editing.

For video generation, Ho~\etal~\citet{hoVideoDiffusionModels2022} have extended the image diffusion approach by using a temporal attention layer for sharing the information along the temporal axis, avoiding costly 3D convolutions.
Other approaches have studied the conditional generation of videos via multiple networks for frame interpolation and super-resolution~\cite{hoImagenVideoHigh2022,singerMakeAVideoTexttoVideoGeneration2022}.
Harvey~\etal~\citet{harveyFlexibleDiffusionModeling2022} implement a diffusion network for generation based on long temporal attention and frame interpolation.
To reduce the computational complexity of handling full 3D videos, multiple 2d projections can be used~\citet{yuVideoProbabilisticDiffusion2023b}.
Mei~\etal~\citet{meiVIDMVideoImplicit2023} have two models: one for frame content and one for frame motion.
The very recent diffusion model Sora~\cite{videoworldsimulators2024} produces videos from text with unprecedented quality. While the complete architecture and training strategy is not known at the time of writing , it is likely that training or even running such a model requires unreasonable computing facilities.

Related to the internal learning strategy proposed in our work, it has also been shown recently that diffusion can be applied in a small data setting, for instance when only a single image is available for training. 
Kulikov~\etal~\citet{kulikovSinDDMSingleImage2023} thus achieve single image generation using only one image. Similarly, Nikankin~\etal~\citet{nikankinSinFusionTrainingDiffusion2022} show the application to image as well as video (single video generation / extrapolation).
It is also possible to finetune a general diffusion model for a specific scene~\cite{ruizDreamBoothFineTuning2023a}.

\subsection{Video Inpainting}

Classical approaches to video inpainting have long relied on available information already present in the video to achieve inpainting of the occluded region. They rely on the \emph{auto-similarity} hypothesis, which is the foundation for patch-based methods in inpainting \cite{wexlerSpaceTimeCompletionVideo2007,newsonVideoInpaintingComplex2014}, as well as other problems such as denoising~\cite{buadesNonlocalAlgorithmImage2005} or image editing~\cite{pritchShiftmapImageEditing2009}.

Another class of approaches is based on \emph{optical flow}. Indeed, in many situations, especially when the occluded region is observed at some point in time in the video, optical flow gives very useful information. \citet{huangTemporallyCoherentCompletion2016} jointly optimize the appearance and the optical flow, making sure that the output is temporally coherent.

More recently, \citet{kimDeepVideoInpainting2019} have proposed a direct inpainting approach using a 3D neural network.
Methods such as~\cite{xuDeepFlowGuidedVideo2019a,gaoFlowedgeGuidedVideo2020} first inpaint the optical flow and then propagate pixel values. These methods have been successful in handling non-rigid motions and long range dependencies.
\citet{zhangInertiaGuidedFlowCompletion2022} proposed to predict the optical flow using more frames, resulting in a more accurate optical flow and better preservation of motions.
\citet{liEndtoEndFrameworkFlowGuided2022a} integrate the optical flow prediction and propagation inside the training loop, optimizing the network parameters for video inpainting end-to-end.

Transformer-based architectures have also been used for video inpainting~\cite{liuFuseFormerFusingFineGrained2021a,zhang2022flow}. Transformers can handle long range dependencies needed in video inpainting, however, their heavy computational cost can be prohibitive for the large dimensionality of videos and imply drastic approximations of the attention.
ProPainter~\cite{zhouProPainterImprovingPropagation2023a} improves the long range propagation of known regions with Transformers.

Internal learning for video inpainting has been proposed by
\citet{zhangInternalLearningApproach2019}. They learn a convolutional model on a single video similarly to Deep Image Prior~\cite{ulyanovDeepImagePrior2018}, and model the optical flow for better temporal coherency.
\citet{ouyangInternalVideoInpainting2021b} tackle the same problem but propose a solution to avoid explicit modeling of the optical flow.
Finally, \citet{renDLFormerDiscreteLatent2022} use a Transformer with a discrete codebook finetuned on the target video, similar to a masked quantized autoencoder.

Only a few works tackle video inpainting using diffusion models, due to their computational requirements. AVID~\cite{zhangAVIDAnyLengthVideo2024}, and CoCoCo~\cite{ziCoCoCoImprovingTextGuided2024} propose different solutions to text-guided video inpainting, which is a different problem than our own \ie automatic video inpainting without user intervention or guidance.
\cite{leeVideoDiffusionModels2024} combine a diffusion model with optical flow for temporal consistency, however the method is still limited when the optical flow is inaccurate.

\section{Method}
\label{sec:method}

\subsection{Background}
\begin{figure*}
    \centering
    \includegraphics[width=\textwidth]{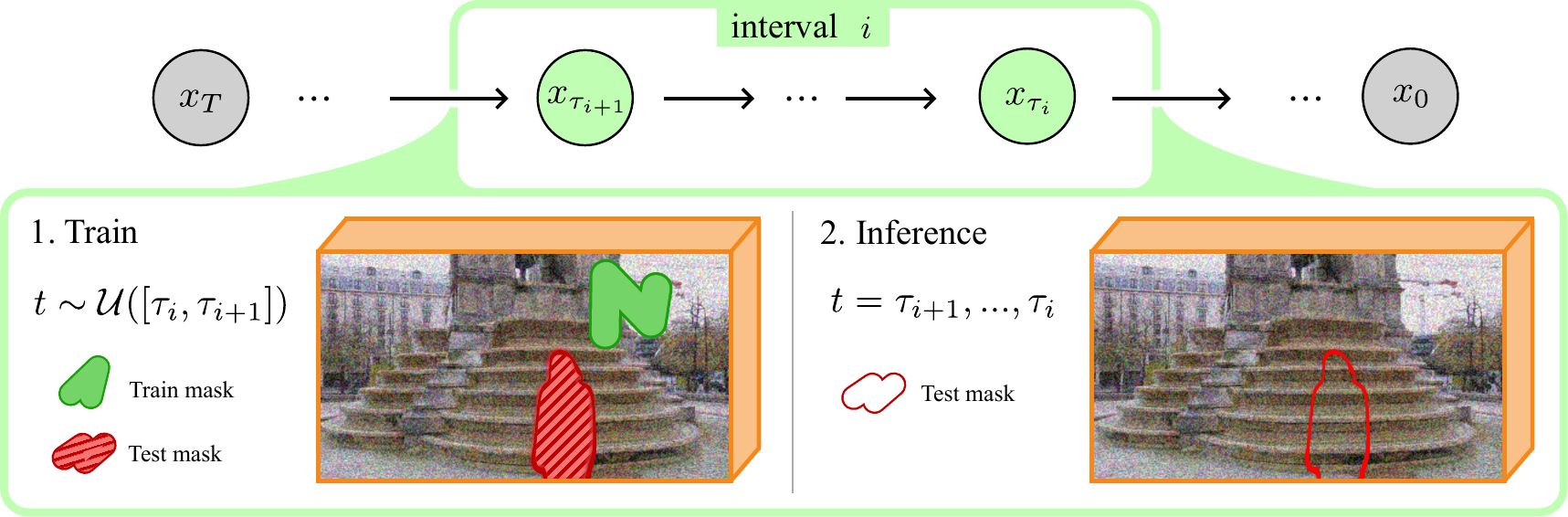}
    \caption{We train a diffusion model on the input video only using interval training. In interval training, we learn on a subset of timesteps and the inference is done immediately after the training phase. Starting from $x_T$, the inpainting result is progressively generated with a perfectly adapted network.}
    \label{fig:sliced}
\end{figure*}

We first recall the diffusion framework of Denoising Diffusion Probabilistic Models (DDPM)~\cite{hoDenoisingDiffusionProbabilistic2020b}, based on discrete forward and backward processes.
We adapt it for inpainting by modeling the conditional data distribution $p(\mathrm{x} \mid y)$ where $\mathrm{x}$ is a full video, and $y$ are the observations \ie the known (unoccluded) region of the video.
We represent an RGB video of $L$ frames of size $H \times W$ with a vector in $\mathbb{R}^{L \times H \times W \times 3}$.
Starting from the unknown distribution of completed videos $q(\mathrm{x}_0 \mid y)$ conditioned on $y$, we define a Markov chain $q(\mathrm{x}_0 \mid y), ..., q(\mathrm{x}_T \mid y)$ with the forward transition kernel for $t \in [1,T]$:

\begin{equation}
    q(\mathrm{x}_{t} \mid \mathrm{x}_{t-1}, y) = \mathcal{N}\left( \mathrm{x}_t; \sqrt{1 - \beta_t} \mathrm{x}_{t-1},  \beta_t \bm{I} \right),
\end{equation}
and the variance schedule linearly increasing from $\beta_1=0.0001$ to $\beta_T=0.02$.
Associated with the forward process is the reverse process defining the data distribution $p_\theta(\mathrm{x}_t \mid \mathrm{x}_{t+1}, y)$ which is also a Gaussian of unknown but learnable mean and variance:

$$
p_\theta \left( \mathrm{x}_t \mid \mathrm{x}_{t+1}, y \right) = \mathcal{N}(\mathrm{x}_t; \mu_\theta(\mathrm{x}_{t+1},y,t), \sigma^2_t\bm{I}).
$$

We use the classical approach of a neural network to predict the conditional mean $\mu_\theta(\mathrm{x}_{t+1}, y, t)$ whose parameters $\theta$ are optimized to maximize
a lower bound %
of the log-likelihood of the data.
Following~\cite{hoDenoisingDiffusionProbabilistic2020b} we use a fixed variance schedule for the reverse process: $\sigma^2_t =  \frac{1 - \bar{\alpha}_{t-1}}{1 - \bar{\alpha}_t} \beta_t$.

In the case of inpainting, we are interested in the conditional data distribution $p(x \mid y)$ where $y$ are our observations \ie the known region of the video, we thus train our network with additional inputs. We minimize the L2 loss of the $x_0$-parameterization~\cite{hoDenoisingDiffusionProbabilistic2020b}:

\begin{equation}
    \mathbb{E}_{\mathrm{x}_0, \epsilon, t, M} \lVert M \circ \left( \mathrm{x}_0 - f_\theta(\tilde{\mathrm{x}}_t, y, t, M) \right) \rVert_2^2,
\end{equation}
where $M$ is a binary mask containing $0$ for the known region and $1$ elsewhere, $y$ is the masked video \ie $y = x_0 \circ (1 - M)$, and $\tilde{\mathrm{x}}_t$ is obtained by the forward diffusion of $\mathrm{x}_0$ by the following:
$$ \tilde{\mathrm{x}}_t = \sqrt{\bar\alpha_t} \mathrm{x}_0 + \sqrt{1 - \bar\alpha_t} \epsilon ,$$
where $\alpha_t = 1-\beta_t$ and $\bar\alpha_t = \Pi_{i=0}^t \alpha_t$, $\epsilon \sim \mathcal{N}(0, I)$.
The observations and masks are introduced via concatenation in the input layer (similarly to \cite{sahariaPaletteImagetoImageDiffusion2022b}).

Unless specified, we use the default parameters of~\cite{hoDenoisingDiffusionProbabilistic2020b}: variance schedule, sampling, time schedule.
Therefore, our method could benefit from many of the recent improvements for each of these items.

\subsection{Architecture}
\label{sec:architecture}

To ensure temporal coherency, we extend the traditional UNet architecture to videos by using 3D convolutions.
Another option is to rely on optical flow, but this can be inaccurate for complex situations such as dynamic textures or complex motion with severe occlusions. 
Another common approach for long-range dependencies is to use the attention mechanism~\cite{vaswani_attention_2017}, but it is extremely computationally heavy, and prohibitively so in the case of videos. Our methods avoids the use of this mechanism, yielding a network with reasonable computational requirements.

Since we are only training the model on a single video, we can greatly reduce the number of parameters in our network in comparison to standard diffusion models.
The network has a total of 16 convolutional layers, distributed over 4 levels. All convolutions are 3x3x3, with padding 1, stride 1. Downsampling is achieved using a max pooling operation while upsampling uses nearest neighbor interpolation. Feature maps from skip connections and feature maps from upsampling layers are combined using concatenation. Our 3D UNet downsamples only in the spatial dimensions to preserve as much as possible the temporal dimension, which is already much smaller than the others. The full architecture details are available in Appendix, Table \ref{tab:unet_architecture}. 
We only use 32 output channels for all internal layers resulting in only 500k parameters which are to be compared to the 500M parameters of Palette~\cite{sahariaPaletteImagetoImageDiffusion2022b} for image inpainting and the billions of parameters of other video diffusion models~\cite{hoImagenVideoHigh2022}.
Our code can be found in the supplementary materials and is available at \url{https://github.com/ncherel/infusion}.

\subsection{Single Video (Internal) Training}
\label{sec:training}

Our neural network is trained to inpaint a single video, using the video itself. This type of approach is referred to as \emph{internal learning}. Fundamentally, it is based on the \emph{auto-similarity} hypothesis, which has been successfully used in the case of patch-based methods~\cite{newsonVideoInpaintingComplex2014} and neural networks via internal learning~\cite{zhangInternalLearningApproach2019,ouyangInternalVideoInpainting2021b}.
The main assumption here, explaining the success of such methods for video inpainting, is that videos contain roughly the same content in successive frames, with some motion.
with some or all of the occluded content being revealed at some point in time. Thus, what might be considered as overfitting in a normal machine learning situation is in fact desirable here: we want to use the content of the video to inpaint itself. Put another way, we take advantage of both the learning and auto-similarity paradigms in our method. This kind of situation also arises in the problem of Single-Image Generation ~\cite{shahamSinGANLearningGenerative2019}.

It is thus possible to efficiently inpaint a video without requiring a large external dataset. We assume that a binary test mask $M_\text{test}$ is provided. It is equal to 1 in the region to be inpainted, 0 elsewhere.
Training is done by sampling consecutive frames from the video, and generating training masks which are to be inpainted / denoised by our neural network.
We prevent data leakage from the test region by additional masking operations for the inputs:

$$
\begin{array}{lcl}
  \tilde{x}_t & = & \sqrt{\bar{\alpha}}_t (1 - M_\text{test}) \circ x_0 + \sqrt{1 - \bar{\alpha}_t} \epsilon \\
  y & = & (1 - M_\text{test}) \circ (1 - M_\text{train}) \circ x_0 \\
  M & = & 1 - (1 - M_\text{test}) \circ (1 - M_\text{train}) \\
\end{array}
$$
$M$ is thus the logical OR combination of $M_\text{test}$ and $M_\text{train}$. Similarly, we ignore some parts of the objective function:
$$
\mathcal{L}_\theta(x_0) = \lVert M_\text{train} \circ (1 - M_\text{test}) \circ \left( x_0 - f_\theta(\tilde{x}_t, y, M, t) \right) \rVert^2
$$

\emph{Internal training vs fine-tuning}. Internal training has the advantage of reducing the complexity of the training problem, but the drawback is that we are in essence overfitting on the video data. Thus, a natural question is whether a generic model can be fine-tuned for the inpainting of a specific video. This could be done using existing video inpainting models~\cite{liEndtoEndFrameworkFlowGuided2022a,zhouProPainterImprovingPropagation2023a} or video generation models~\cite{hoVideoDiffusionModels2022}. We carried out experiments to explore these options, and found that neither options were immediately promising. Fine-tuning a video inpainting model in fact lead to worse perceptual metrics. The large generative diffusion model generated unconvincing results in the inpainting setting. We refer the reader to the appendix for details.

\subsection{Interval training}
\label{sec:schedule}

Training a diffusion network obviously involves training it for all timesteps $t$, from $1$ to $T$, since all these steps are needed during inference.
However, it is clear that, depending on the timestep, the goal of the network, to estimate $x_0$ will be very different: if $t$ is large (pure noise), then the network should behave more like a generative model. If $t$ is small, then the network should act more like a denoiser, since the original video is only slightly noisy. This observation is crucial: indeed, in a standard diffusion model, one single network is supposed to achieve these very different goals. This is one reason why diffusion networks tend to be extremely large.

To circumvent this problem, we propose to divide up the timesteps into subsets, which we call \emph{intervals}. We use one lightweight network, trained on one interval at a time. The training starts from the end of the diffusion process on a first interval.
Once training is done on this interval, we use the model to perform the inference on the test data for all timesteps of the interval, up to the beginning of the next interval. The model is then trained on the next time interval. This is carried out until we have reached time step $t=0$. We can, of course, save the intermediate models if we wish, but this is not really necessary for video inpainting, since the model can only be used for the current video.

In practice, we define a time schedule of $N$ steps such that $\tau_1 = 1 < \tau_1 < ... <  \tau_N = T$.
Our network learns successively on each interval $[\tau_{i}, \tau_{i+1}]$, starting from the most noisy interval.
After training on an interval, the inference is done for the same interval \ie for all timesteps from $\tau_{i+1}$ to $\tau_{i}$.
Training is detailed in Figure~\ref{fig:sliced} and Alg. \ref{alg:training}.

\begin{algorithm}[t]
    \caption{Training / inference algorithm for interval training}
    \begin{algorithmic}
        \Require{model $f_\theta$, intervals $\tau_{i=1..N}$, test data $y_{\text{test}}$}
        \State $x_{\text{test}} \gets \mathcal{N}(0, I)$
        \For{$i = N-1, ..., 1$}
            \For{$j = 1,.., K$} \Comment{Training for $K$ iterations}
                \State $x_j, y_j, \epsilon_j, M_j \gets \text{NextBatch}(j)$
                \State $t \gets \mathcal{U}(\tau_i, \tau_{i+1})$
                \State Take gradient step with gradient:
                \State $\quad \nabla_\theta \lVert x_j - f_\theta(\sqrt{\bar\alpha_t} x_j + \sqrt{1 - \bar\alpha_t} \epsilon_j, y_j, t, M_j)\rVert^2$
            \EndFor

            \For{$t = \tau_{i+1}, ..., \tau_{i}$} \Comment{Inference}
                \State $\mu_\theta \gets f_\theta(x_{\text{test}}, y_{\text{test}}, t)$
                \State $x_{\text{test}} \sim \mathcal{N}(\mu_\theta, \sigma^2_t I)$
            \EndFor
        \EndFor
        \State \Return $x_{\text{test}}$
      \end{algorithmic}
    \label{alg:training}
\end{algorithm}

\begin{table*}
    \centering
    \caption{Inpainting metrics on dynamic texture datasets. 
    }
    \begin{tabular}{|c|c|c|c|c|c|c|c|c|c|}
      \hline
      \multirow{ 2}{*}{Method} & \multicolumn{5}{c|}{Dynamic textures (Tesfaldet)} & \multicolumn{4}{c|}{Dynamic textures (DTDB)} \\
      \cline{2-10}
 & LPIPS ↓ & SVFID ↓ & PSNR ↑ & SSIM ↑ & Diversity ↑ & LPIPS ↓ & SVFID ↓  & PSNR ↑ & SSIM ↑\\
\hline
Patch~\cite{newsonVideoInpaintingComplex2014} & 0.051 & 0.227 & 29.47 & 0.954 & 0.053 & 0.0180 & 0.055  & 36.52 & 0.988\\
ISVI~\cite{zhangInertiaGuidedFlowCompletion2022} & 0.062 & 0.500 & 27.41 & 0.945 & 0 & 0.0302 & 0.248 & 34.20 & 0.982\\
E\textsuperscript{2}FGVI~\cite{liEndtoEndFrameworkFlowGuided2022a} & 0.046 & 0.288 & 31.58 & 0.961 & 0 & 0.0145 & 0.078   & 38.76 & \textbf{0.991}\\
FGT~\cite{zhangFlowGuidedTransformerVideo2022} & 0.046 & 0.202 & 30.88 & 0.956 & 0 & 0.0144 & 0.059  & 37.99 & 0.990\\
ProPainter~\cite{zhouProPainterImprovingPropagation2023a} & 0.046 & 0.261 & \textbf{31.77} & 0.960 & 0 & \textbf{0.0137} & 0.051 & \textbf{39.62} & \textbf{0.991}\\
\hline
Infusion (ours) & \textbf{0.035} & \textbf{0.116} & 30.27 & \textbf{0.964} & \textbf{0.054} & \textbf{0.0137} & \textbf{0.036} & 37.20 & 0.989\\
        \hline
    \end{tabular}
    \label{tab:textures}
\end{table*}

This divide-and-conquer strategy has several advantages. Firstly, it allows us to achieve high quality video inpainting performance with a lightweight diffusion network, since each network specializes only one timestep interval. Secondly, we can avoid complicated weighting schemes for the loss function: in one interval, all timesteps have roughly the same weight and the weighting term can thus be set to $1$. This is not the case when one network is used and must perform well on all $T$ timesteps~\cite{hangEfficientDiffusionTraining2023,nicholImprovedDenoisingDiffusion2021a}.

\section{Experiments}
\label{sec:experiments}

\subsection{Evaluation and masks}

We compare our method with several state of the art methods. We include a classical patch-based approach \cite{newsonVideoInpaintingComplex2014} which is also internal and well adapted to the inpainting of dynamic textures, as well as more recent deep learning approaches: \EFGVI~\cite{liEndtoEndFrameworkFlowGuided2022a}, ISVI~\cite{zhangInertiaGuidedFlowCompletion2022}, FGT~\cite{zhangFlowGuidedTransformerVideo2022}, and ProPainter~\cite{zhouProPainterImprovingPropagation2023a}.
Compared to our approach, these methods are not trained or fine-tuned on the test video, but this does not always bring better performance. Details and discussion can be found in the appendix.

We evaluate our method on complex videos featuring dynamic textures, large occlusions, and moving objects. These include, firstly, video sequences with a segmented object to be removed, from \cite{newsonVideoInpaintingComplex2014,granadosHowNotBe2012,leAGM19}, as well as some new sequences also with segmented objects. Secondly, we use dynamic texture databases: \cite{tesfaldetTwoStreamConvolutionalNetworks2018}, featuring 59 textures, and a subset of DTDB~\cite{hadjiNewLargeScale2018} containing one random video from each of the 45 appearance categories. 
In these last two databases, we use synthetic masks, since there are no meaningful objects (people walking, etc) to be removed.

We evaluate our method qualitatively on the first set of sequences, with objects to be removed, for which no ground truth is available. We evaluate our method quantitatively on the second set of sequences, using various classical and perceptual metrics.

\begin{figure*}
    \centering
    \begin{tabular}{ccc}
    \includegraphics[width=0.3\textwidth]{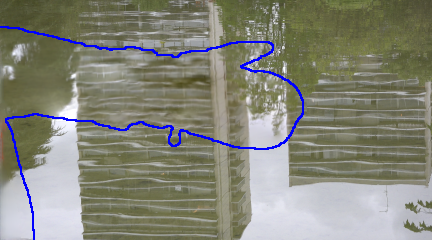}
    & \includegraphics[width=0.3\textwidth]{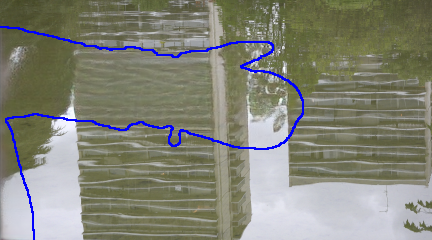}
    & \includegraphics[width=0.3\textwidth]{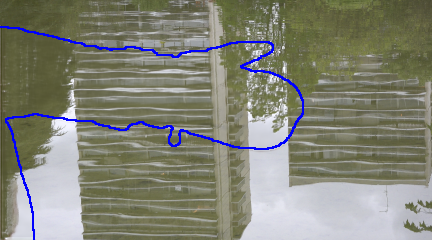} \\

    \includegraphics[width=0.3\textwidth]{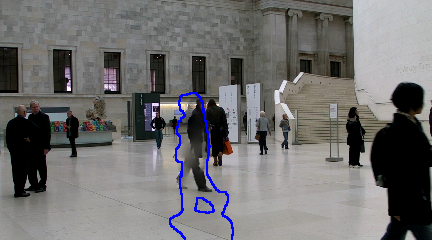}
    & \includegraphics[width=0.3\textwidth]{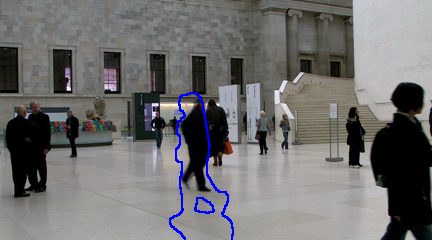}
    & \includegraphics[width=0.3\textwidth]{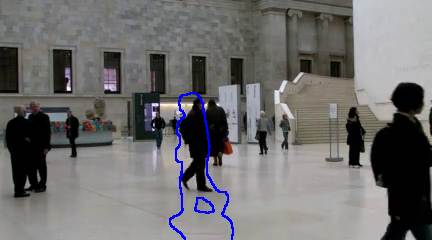} \\
    
    \includegraphics[width=0.3\textwidth]{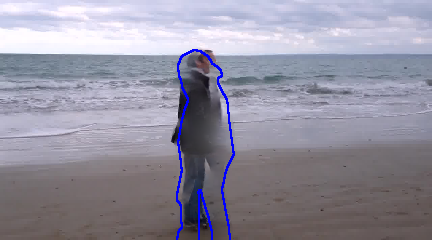}
    & \includegraphics[width=0.3\textwidth]{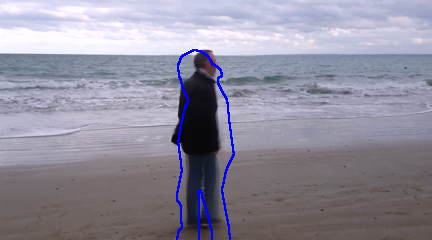}
    & \includegraphics[width=0.3\textwidth]{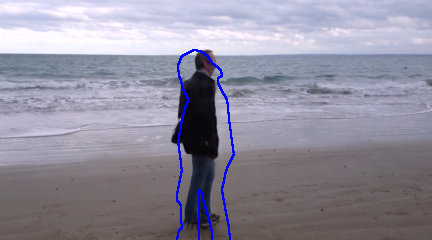} \\

    FGT & ProPainter & Infusion
    \end{tabular}
    
    \caption{In these scenes, our method satisfactorily inpaints dynamic textures preserving sharp details (top). It also better handles complex occlusions such as people crossing paths.}
    \label{fig:occlusion}
\end{figure*}

\subsection{Training details}

We resize the training data and masks to 432x240 for sequences larger than that.
While our architecture is lightweight and may thus be used with many frames on a reasonable GPU, we only use 20 frames at the same time for training.
Contrary to other approaches for video generation, we use all consecutive frames without dropping any of them.

For training, we generate masks using random strokes similarly to \citet{kimDeepVideoInpainting2019}. Training masks may overlap with the test masks, in which case no loss is actually computed in the intersection.

We use the Adam optimizer with an initial learning rate of 1e-4. We train for 200k steps for object removal, this takes 15 hours. For dynamic texture inpainting which is of lower time and spatial resolution, we train for 40k steps (3 hours).
For interval training, the 1000 timesteps are divided into 20 intervals of length 50. The same training budget is allocated to each interval.

\subsection{Quantitative results}
\label{sec:results}

We quantitatively evaluate the methods using Peak Signal-to-Noise Ratio (PSNR), structure similarity measure (SSIM), Learned Perceptual Image Patch Similarity~\cite{zhangUnreasonableEffectivenessDeep2018a} (LPIPS) and Single Video Fréchet Inception Distance (SVFID)~\cite{wangVideotoVideoSynthesis2018,gurHierarchicalPatchVAEGAN2020}.
Finally, we also compute a diversity score following \cite{shahamSinGANLearningGenerative2019}. As explained above, the evaluation is conducted on the two databases of dynamic textures. 

The first three metrics (PSNR, SSIM and LPIPS)  are computed frame-by-frame and therefore do not reflect motion discontinuities. 
SVFID, on the other hand, takes spatio-temporal features into account.
For computing the SVFID, we follow \cite{dorkenwaldStochasticImagetoVideoSynthesis2021} and use the same model, trained on dynamic textures, for extracting the spatio-temporal features required for its computation.

Results are given in Table~\ref{tab:textures}. We first observe that the LPIPS metric is significantly better with our method. In contrast, PSNR and SSIM are relatively good for all methods, the SSIM being slightly better for us and the PSNR better for the two deep-learning based methods \cite{zhouProPainterImprovingPropagation2023a}and \cite{liEndtoEndFrameworkFlowGuided2022a}. A visual inspection of the videos provided with this paper nevertheless clearly shows that such methods struggle to reproduce spatial details of the textures. This can be explained by the fact that the PSNR is well-known to be poorly adapted to fine details \cite{blauPerceptionDistortionTradeoff2018,sahariaImageSuperResolutionIterative2022}. The LPIPS metric, on the other hand, is more faithful to human perception.

We also observe on Table~\ref{tab:textures} that the SVFID metric, the only truly video-based metric considered here, is significantly better for our approach. This shows that our method respects spatio-temporal statistics better than competing methods. In fact, a visual inspection of the resulting video (included in the supplementary materials) shows that for such content, deep-learning based competing approaches (ISVI, FGT, \EFGVI, and ProPainter) often produce results that are almost still or have completely wrong dynamics. This behavior is likely to be the result of a heavy use of optical flow to extend two-dimensional deep learning-based methods to videos. In addition, these methods have been trained on general purpose video databases, in which dynamic textures are often poorly represented. In contrast, the internal training strategy of our method and the fact that it is based on the diffusion framework with a purely convolutional network (no optical flow) is especially adapted to dynamic textures and high spatial and temporal frequency.  

The diversity score shows better results of our approach and of the patch-based method from \cite{newsonVideoInpaintingComplex2014} with respect to other approaches. In fact, all considered deep-learning methods always produce the same result, yielding a null diversity score.

The diversity of our algorithm outcomes also enables us to perform the following experiment. For each texture of the Tesfaldet dataset, we produce 100 outcomes and average them to produce a new result, included in Table~\ref{tab:average} as \textit{infusion (avg)}. This averaged result has a significantly better PSNR score than single outcomes. As expected, this improvement in PSNR comes at the cost of poorer fine details (Figure~\ref{fig:average} and videos in the supplementary materials) and lower perceptual and diversity scores, more in line with the performance of the best methods among baselines \EFGVI, ISVI, FGT, and ProPainter. This is not surprising since these methods are not generative models but regressors. We believe that this simple experiment cleary higlights the strong limitation of the PSNR to evaluate results, in particular in presence of high frequency content.

\begin{table}
    \centering
    \caption{Ablation study on the interval length during training/inference. Using no interval (length 1000) is significantly worse than our chosen interval size (50).}
    \begin{tabular}{|c|c|c|c|c|}
      \hline
      \multirow{ 2}{*}{Interval length} & \multicolumn{4}{c|}{Dynamic textures (Tesfaldet)} \\
     \cline{2-5}
        & LPIPS ↓ & SVFID ↓ & PSNR ↑ & SSIM ↑ \\
       \hline
        1000  & 0.042 & 0.145 & 29.54 & 0.960 \\
        100 & 0.042 & 0.121 & 29.78 & 0.957\\
        50 & \textbf{0.035} & 0.116 & \textbf{30.27} & \textbf{0.964}\\
        10 & 0.039 & \textbf{0.111} & 30.12 & 0.958 \\
        1 & 0.043 & 0.137 & 30.10 & 0.958 \\
        \hline
    \end{tabular}
    \label{tab:ablation}
\end{table}

\begin{table}
    \centering
    \caption{Inpainting metrics when averaging 100 samples or using a single one on the Tesfaldet dataset. PSNR is improved by averaging but the perceptual metrics are severely degraded.}
    \begin{tabular}{|c|c|c|c|c|}
      \hline
        Method & LPIPS ↓ & SVFID ↓ & PSNR ↑ & SSIM ↑\\
        \hline
        Ours (avg) & 0.056 & 0.231 & \textbf{30.97} & 0.962\\
        Ours & \textbf{0.035} & \textbf{0.116} & 30.27 & \textbf{0.964} \\
        \hline
    \end{tabular}
    \label{tab:average}
\end{table}

\begin{table}
    \centering
    \caption{Inference times, for a 432x240 video on an NVidia V100.}
    \begin{tabular}{|c|c|}
    \hline
    Method & \makecell{Inference speed\\(second / frame)} \\
      \hline
      Patch~\cite{newsonVideoInpaintingComplex2014} & 5.13\\
ISVI~\cite{zhangInertiaGuidedFlowCompletion2022} & 1.59 \\
\EFGVI~\cite{liEndtoEndFrameworkFlowGuided2022a} & 0.10 \\
FGT~\cite{zhangFlowGuidedTransformerVideo2022} & 1.83 \\
ProPainter~\cite{zhouProPainterImprovingPropagation2023a} & 0.14 \\
Infusion (ours) & 5.07 \\
        \hline
    \end{tabular}
    \label{tab:time}
\end{table}

\subsection{Object removal experiments}

In this section, we consider the problem of object removal through several classical sequences used in previous works on video inpainting \cite{newsonVideoInpaintingComplex2014,granadosHowNotBe2012,leAGM19} and a few new ones. We consider the same methods as before. For each sequence, a mask is given corresponding to an object (often a person) to be removed. As already mentioned, there is no ground truth for these sequences, so that quantitative evaluation is not performed. Nevertheless, such situations correspond to real applications and a more realistic foreground-background articulation than with synthetic masks. All results can be found in the supplementary materials. One can observe a much better respect of backgrounds such as dynamic textures or occluded moving objects with our approach than with other deep-learning based ones. This is illustrated in Figure~\ref{fig:occlusion}. On the first line, one sees that the unstable reflection of the building in the water is better reproduced by our methods than by FGT and ProPainter. The second and third rows display cases with occluded moving objects. In this case, results are blurry for ProPainter and blending artifacts are visible for FGT, while ours yields more satisfactory structures. We encourage the reader to view the videos in the supplementary materials. These videos support the claim that a diffusion framework with appropriate internal learning is a good choice for stochastic and complex content. 
Our method can however suffer from some artifacts as seen in the \textit{loulous} sequence (see supplementary materials). A black artifact is present when the inpainting mask overlaps the video boundaries in the last frames, this is due to our simplistic (training) mask generation technique.

\subsection{Training and inference times}
\label{sec:training}

Inference speeds in second / frame for all methods can be found in Table~\ref{tab:time}. We can see that our method is not the fastest, performing on par with the patch-based method of Newson~\etal. This is a known limitation of diffusion models with a large number of diffusion steps.

For training times, we cannot compare with the competing methods because the relevant information is not always available or is valid for a different hardware configuration.
For our method on the dynamic texture datasets, training takes 3 hours on average for the Tesfaldet dataset and 6 hours on average for the DTDB dataset which has a higher resolution.
All trainings are done using one NVidia V100.
For object removal, we increase the number of training iterations, which results in a training time of around 15 hours.
Training times could possibly be reduced by adopting latent diffusion models~\cite{rombachHighResolutionImageSynthesis2022b}.

\subsection{Ablation study}
\label{sec:ablation}

\begin{figure}
  \centering
  \setlength{\tabcolsep}{1.5pt}
  \begin{tabular}{cc}
    \includegraphics[width=0.22\textwidth]{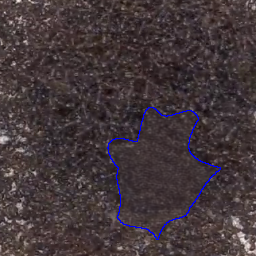} & \includegraphics[width=0.22\textwidth]{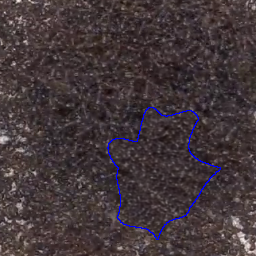} \\
    \includegraphics[width=0.22\textwidth]{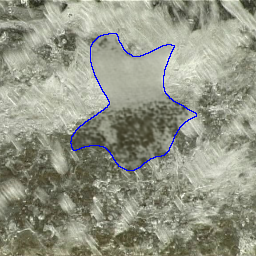} & \includegraphics[width=0.22\textwidth]{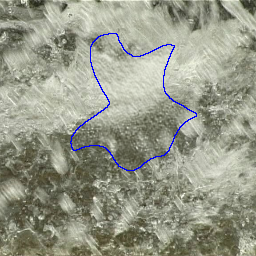} \\
   Baseline training & Interval training
    \end{tabular}
    \caption{Interval training clearly improves the visual results, especially the finer details. Associated videos are found in the supplementary materials.}
    \label{fig:slice}
\end{figure}

This section demonstrates the importance of interval training. We compare the baseline training (\ie training on all timesteps) and interval training.
We train a neural network with the same number of parameters and for the same total number of training steps for both approaches.
Two visual results can be seen in Figure~\ref{fig:slice}.  We observe that interval training clearly improves the reproduction of fine, textured details. More results can be seen on the videos in the supplementary materials.

We also evaluate the impact of the interval length size quantitatively.
Results can be found in Table~\ref{tab:ablation} and indicates that there is an optimal interval length for $l=50$ which is the default parameter we use in the experiments.
Note that all these experiments are done with the same total training budget.
Interval training significantly improves the quantitative results, especially for the perceptual metrics (LPIPS, SVFID).

We recall that in the baseline approach, results are very sensitive to the weighting scheme of the loss function~\cite{nicholImprovedDenoisingDiffusion2021a,hangEfficientDiffusionTraining2023}. Our interval training approach has no such weighting, since each interval is trained separately, and uniform weighting on the interval is sufficient in all cases. Thus, our algorithm does not require time-consuming tuning of the loss weighting, which is a significant advantage.

\subsection{Limitations and failure cases}

\begin{table}
    \centering
    \caption{Inpainting metrics on the DAVIS dataset evaluation set. }
    \begin{tabular}{|c|c|c|}
      \hline
        Method & PSNR (dB) ↑ & SSIM ↑\\
        \hline
        ISVI~\cite{zhangInertiaGuidedFlowCompletion2022} & 30.34 & 0.946\\
        E\textsuperscript{2}FGVI~\cite{liEndtoEndFrameworkFlowGuided2022a} & 33.71 & 0.970 \\
        ProPainter~\cite{zhouProPainterImprovingPropagation2023a} & \textbf{34.43} & 0.973 \\
        Infusion (ours) & 31.55 & \textbf{0.975} \\
        \hline
    \end{tabular}
    \label{tab:davis}
\end{table}

\begin{figure}
  \centering
  \setlength{\tabcolsep}{1.5pt}
  \begin{tabular}{cc}
    \includegraphics[width=0.22\textwidth]{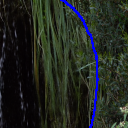} & \includegraphics[width=0.22\textwidth]{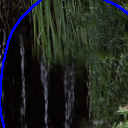} \\
    Frame \#65 & Frame \#99
  \end{tabular}
  \caption{When generating content, our method may not guarantee long term temporal consistency, especially for textures.}
  \label{fig:failure}
\end{figure}

Our method relies on a convolutional architecture which has a limited receptive field, especially in the temporal dimension.
Our method may have trouble reproducing still background for long periods of time. This particularly degrades the performance for static background reconstruction as shown by the comparison with other methods on the DAVIS dataset (Table~\ref{tab:davis}). On this dataset, which contains mostly still background, the positioning of our method is less favorable compared to other methods.
This limitation is most visible for regions that are never seen in the video.
We show an example in Figure~\ref{fig:failure} where the region inpainted changes from grass to water over a long occlusion (more than 30 frames apart) when it should stay stable.

Another limitation of our method is that each model is ``single-use'', in the sense that it is adapted to a single video. This makes sense in the case of video inpainting, and greatly reduces network size and training time but it means that the model cannot be re-used, as it can be in other approaches. The optical flow is more generalizable from one video to the next but, as we have seen, is limiting for elements such as dynamic textures.

Finally, our method is fully internal and thus relies on content from the video itself. Thus, it may have difficulty introducing completely new content in the video.
An example of this situation can be seen in the Appendix where an object only half-occluded is not semantically inpainted.
Adding such completely new content, while maintaining a reasonable network size is a significant challenge for future research.

\section{Conclusion}

We have proposed a diffusion-based method for video-inpainting. 
This is made possible thanks to our novel training scheme and a lightweight network (around 500k parameters vs several billions for most video synthesis methods using diffusion). In particular, our training method leverages only the information contained in the current video, which is sufficient to achieve high quality inpainting, as confirmed by very satisfying results both quantitatively and qualitatively. We show a dramatic improvement for dynamic textures over current methods.

A future research direction would be to handle all situations, \ie extreme camera motion and dynamic/complex scenes, within the same method.
A second direction would be to accelerate training and inference times. To do so we can draw inspiration from Latent Diffusion Models~\cite{rombachHighResolutionImageSynthesis2022b}, distillation techniques~\cite{Salimans2022}, or multi-scale diffusion models~\cite{Zhou2023}. %
Finally a major challenge is to extend the method to situations where the inpainted region should contain an element that does not resemble anything visible in the non-occluded part of the sequence. 

\section*{Acknowledgments}
This study has been carried out with partial financial support
from the French Research Agency through the projects PostProdLEAP
(ANR-19-CE23-0027-01), MISTIC (ANR-19-CE40-005), and IDeGeN (ANR-21-CE23-0024).
Nicolas Cherel is supported by the Institut Mines Télécom, Fondation Mines-Télécom, and l’Institut Carnot TSN through the ``Futur et Ruptures'' grant.

\begin{figure*}
\setlength{\tabcolsep}{1.5pt}
\centering
\begin{tabular}{cccc}
    \includegraphics[width=0.21\textwidth]{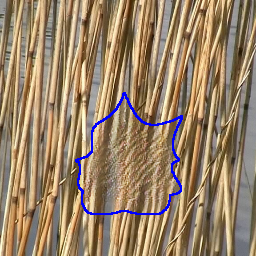}
    & \includegraphics[width=0.21\textwidth]{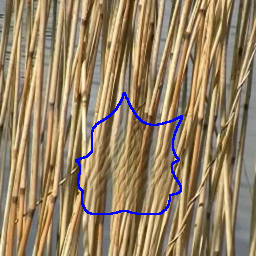}
    & \includegraphics[width=0.21\textwidth]{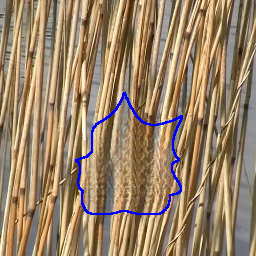}
    & \includegraphics[width=0.21\textwidth]{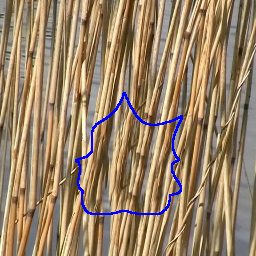} \\

    \includegraphics[width=0.21\textwidth]{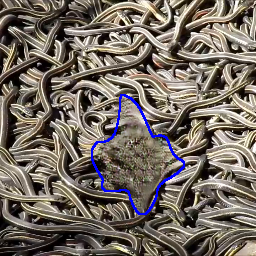}
    & \includegraphics[width=0.21\textwidth]{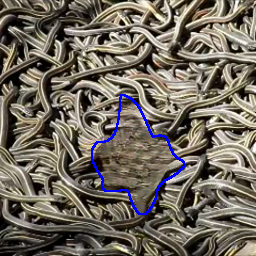}
    & \includegraphics[width=0.21\textwidth]{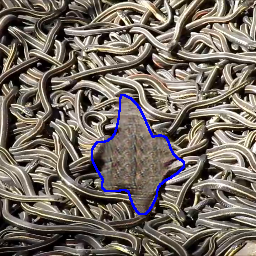}
    & \includegraphics[width=0.21\textwidth]{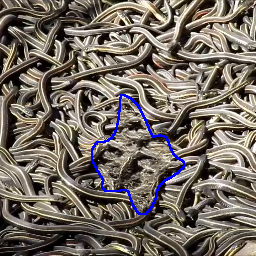} \\

    \includegraphics[width=0.21\textwidth]{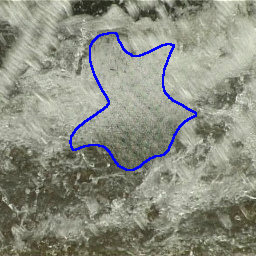}
    & \includegraphics[width=0.21\textwidth]{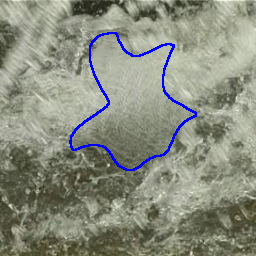}
    & \includegraphics[width=0.21\textwidth]{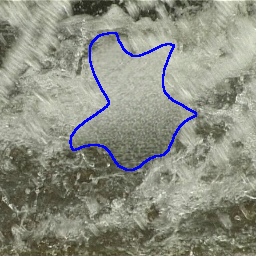}
    & \includegraphics[width=0.21\textwidth]{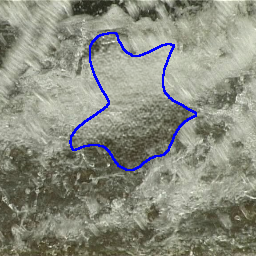} \\
    FGT & \EFGVI & ProPainter & Infusion \\
\end{tabular}
\caption{On all these challenging textures, our method better inpaints the missing region. The motions are also much better inpainted (see video in the supplementary materials).}
\label{fig:textures}
\end{figure*}

\begin{figure*}
    \centering
    \begin{tabular}{ccc}
    \includegraphics[width=0.3\textwidth]{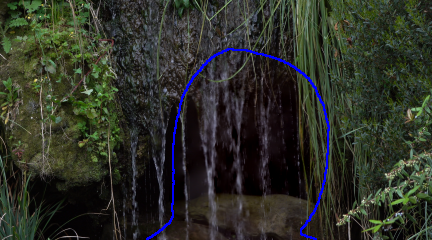}
    & \includegraphics[width=0.3\textwidth]{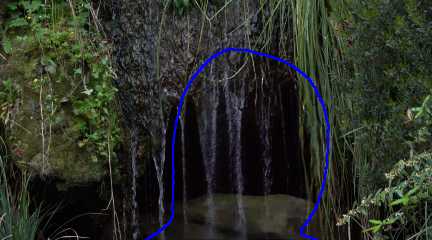}
    & \includegraphics[width=0.3\textwidth]{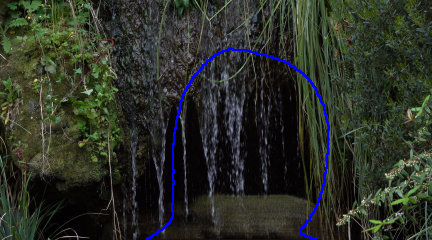} \\

    \includegraphics[width=0.3\textwidth]{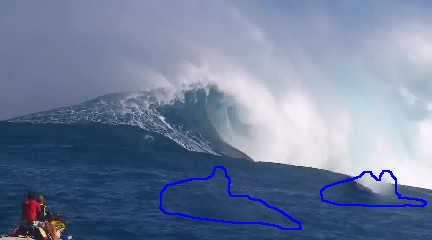}
    &\includegraphics[width=0.3\textwidth]{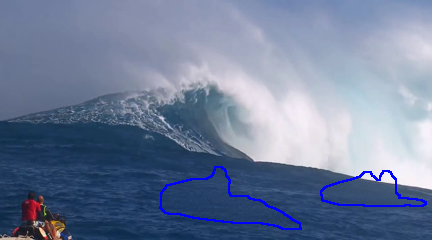}
    &\includegraphics[width=0.3\textwidth]{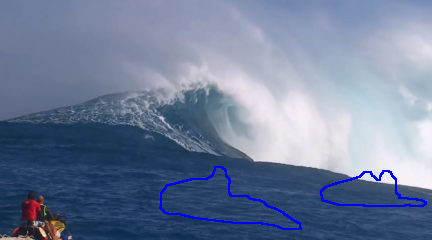} \\
    FGT & ProPainter & Infusion
    \end{tabular}
    
    \caption{In these scenes, we already see in static images that our method significantly outperforms other approaches which are either blurry or contain network artifacts. The result is even more clear when played as a video as optical-flow based methods are unable to reproduce the convincing dynamics of the textures.}
    \label{fig:object_textures}
\end{figure*}

\begin{figure*}
\centering
    \raisebox{0.055\textwidth}{\rotatebox[origin=c]{90}{\EFGVI}}
    \includegraphics[width=0.23\textwidth]{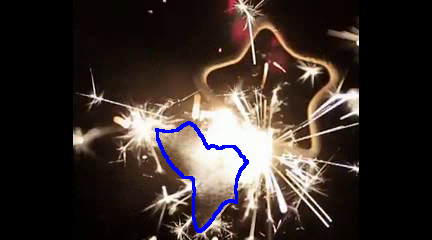}
    \includegraphics[width=0.23\textwidth]{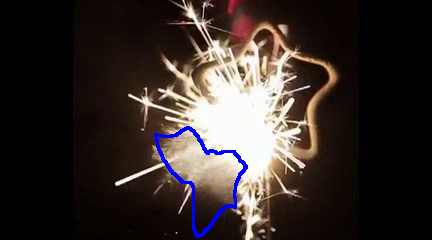}
    \includegraphics[width=0.23\textwidth]{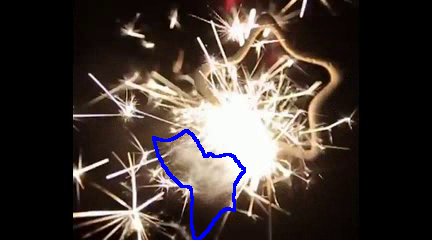}
    \includegraphics[width=0.23\textwidth]{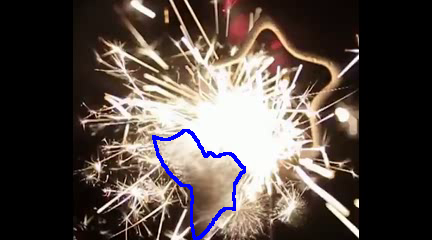}

    \raisebox{0.055\textwidth}{\rotatebox[origin=c]{90}{Infusion}}
    \includegraphics[width=0.23\textwidth]{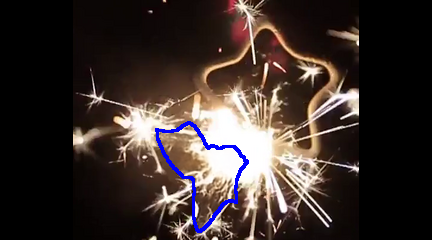}
    \includegraphics[width=0.23\textwidth]{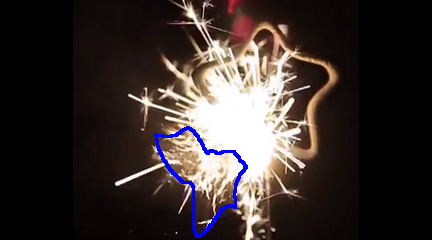}
    \includegraphics[width=0.23\textwidth]{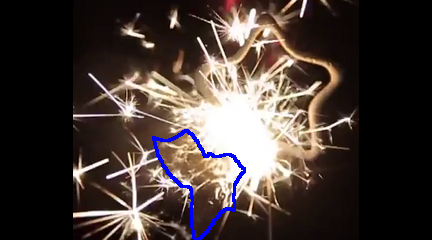}
    \includegraphics[width=0.23\textwidth]{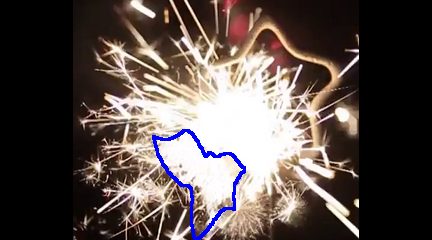}

    \raisebox{0.055\textwidth}{\rotatebox[origin=c]{90}{FGT}}
    \includegraphics[width=0.23\textwidth]{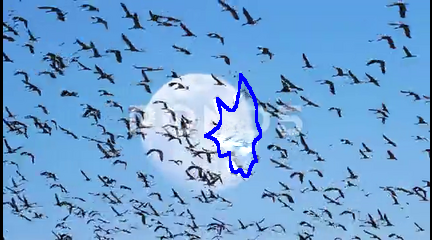}
    \includegraphics[width=0.23\textwidth]{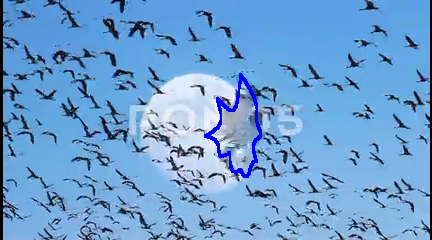}
    \includegraphics[width=0.23\textwidth]{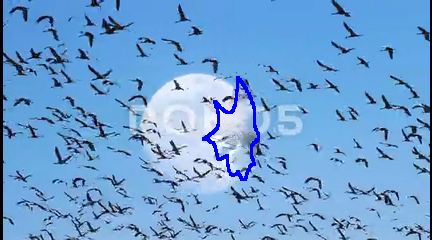}
    \includegraphics[width=0.23\textwidth]{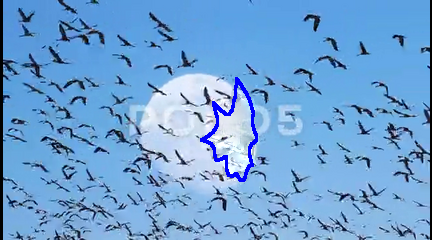}

    \raisebox{0.055\textwidth}{\rotatebox[origin=c]{90}{Infusion}}
    \includegraphics[width=0.23\textwidth]{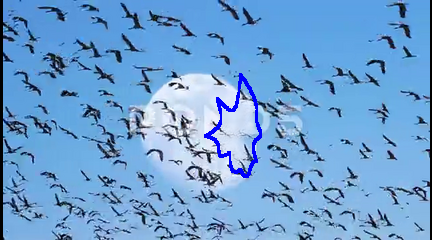}
    \includegraphics[width=0.23\textwidth]{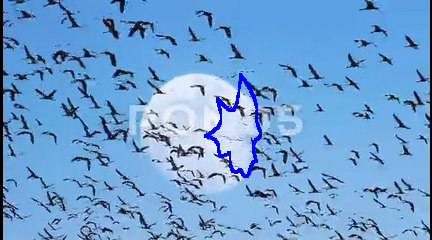}
    \includegraphics[width=0.23\textwidth]{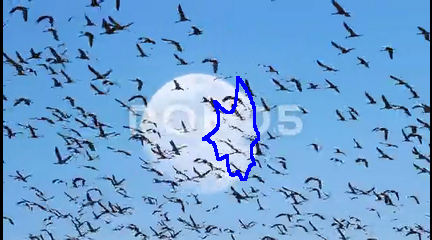}
    \includegraphics[width=0.23\textwidth]{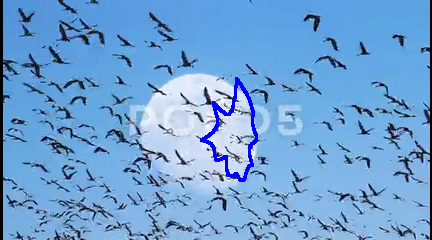}

    \caption{Examples from DTDB. Our method is able to inpaint highly dynamic textures which is a current limitation of flow-based methods like \EFGVI, or FGT.}
\end{figure*}

\begin{figure*}
\centering
\begin{tabular}{ccc}
    \includegraphics[width=0.26\textwidth]{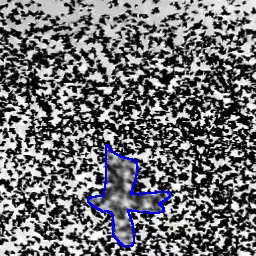}
    & \includegraphics[width=0.26\textwidth]{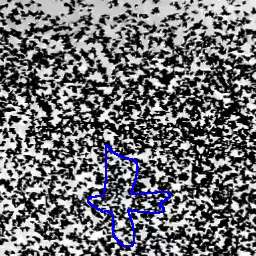}
    & \includegraphics[width=0.26\textwidth]{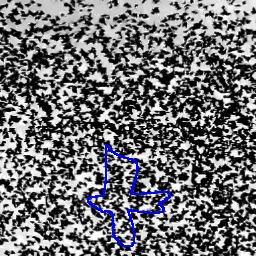} \\
    Average of 100 samples & Sample \# 1 & \# Sample 2
    \end{tabular}
    \caption{Average of 100 diffusion samples vs independent samples. Averaging leads to a better PSNR (22.78 dB vs 20.66 dB vs 20.60 dB) but degrades the results visually. For stochastic textures, the PSNR is not a relevant metric.}
    \label{fig:average}
\end{figure*}

\bibliographystyle{eg-alpha-doi} 
\bibliography{references}

\newpage
\appendix

\section{Fine-tuning on test videos}

Fine-tuning is now a common practice with the availability of large foundation models~\cite{rombachHighResolutionImageSynthesis2022b} and efficient techniques such as LoRA~\cite{hu2022lora}.
While fine-tuning sounds more reasonable than training a model from scratch for each video, we show in this section that this does not always generate good results.

First, we show that fine-tuning does not bring additional performance for ProPainter~\cite{zhouProPainterImprovingPropagation2023a}.
ProPainter indeed is trained on a separate external dataset and could benefit from a refinement step on a specific video.
However we see in Table~\ref{tab:finetuning}), that training ProPainter for an additional 200 steps (20 min) improves the PSNR but degrades the perceptual metrics such as LPIPS and SVFID.
Note that ProPainter could only be fine-tuned using the L2 loss and not the discriminator loss. The discriminator cannot be used when only one partially masked video is available as training data.

\begin{table}
    \centering
    \caption{Inpainting results after fine-tuning on the Tesfaldet texture dataset. Fine-tuning ProPainter on the test video degrades the perceptual metrics (LPIPS, SVFID) but improves the PSNR. FT=fine-tuned}
    \begin{tabular}{|c|c|c|c|c|}
    \hline
        Method & LPIPS ↓ & SVFID ↓ & PSNR↑ & SSIM ↑  \\ \hline
        Ours & \textbf{0.035} & \textbf{0.116} & 30.27 & \textbf{0.964} \\ \hline
        ProPainter & 0.046 & 0.261 & 31.77 & 0.960 \\ \hline
        ProPainter FT* & 0.053 & 0.367 & \textbf{32.11} & 0.962 \\ \hline
    \end{tabular}
    \label{tab:finetuning}
\end{table}

Second, we tried to use one of the few video diffusion models available: Stable Video Diffusion~\cite{blattmann2023stablevideodiffusionscaling}.
Stable Video is however not designed to do video inpainting, so several adjustments have to be made. Specifically, we adopted the approach of Lugmayr~\etal in RePaint~\cite{Lugmayr_2022_CVPR} by removing the conditioning information and performing several time travels while reprojecting the visible regions onto the generated video.
Unfortunately, this lead to poor results with unrealistic completions lacking coherence with the visible part of the video.

\begin{table}[h]
\centering
\caption{U-Net Architecture Details (4 levels) - All convolutions are 3x3x3 and use ReLU.}
\begin{tabular}{|l|l|l|l|}
\hline
Layer & Type & \makecell{Output Shape \\ (T, H, W, C)} & Extra \\
\hline
Input & Concat & (T, 256, 256, 23)  & $\tilde{x}_t, y, t, M$ \\
Conv1\_0 & Conv3D & (T, 256, 256, 32) & - \\
Conv1\_1 & Conv3D & (T, 256, 256, 32)  & - \\
Conv1\_2 & Conv3D & (T, 256, 256, 32) & $\rightarrow$ Skip1 \\
Pool1 & MaxPool2D & (T, 128, 128, 32) & - \\
Conv2\_1 & Conv3D & (T, 128, 128, 32) & - \\
Conv2\_2 & Conv3D & (T, 128, 128, 32) & $\rightarrow$ Skip2 \\
Pool2 & MaxPool2D & (T, 64, 64, 32) & - \\
Conv3\_1 & Conv3D & (T, 64, 64, 32) & - \\
Conv3\_2 & Conv3D & (T, 64, 64, 32) & $\rightarrow$ Skip3 \\
Pool3 & MaxPool2D & (T, 32, 32, 32) & - \\
Conv4\_1 & Conv3D & (T, 32, 32, 32) & - \\
Conv4\_2 & Conv3D & (T, 32, 32, 32) & - \\
Up3 & Upsample & (T, 64, 64, 32) & - \\
Concat & Concat & (T, 64, 64, 64) & Up3, Skip3\\
Conv5\_1 & Conv3D & (T, 64, 64, 32) & - \\
Conv5\_2 & Conv3D & (T, 64, 64, 32) & - \\
Up2 & Upsample & (T, 128, 128, 32) & - \\
Concat & Concat & (T, 128, 128, 64) & Up2, Skip2\\
Conv6\_1 & Conv3D & (T, 128, 128, 32) & - \\
Conv6\_2 & Conv3D & (T, 128, 128, 32) & - \\
Up1 & Upsample & (T, 256, 256, 32) & - \\
Concat & Concat & (T, 256, 256, 64) & Up1, Skip1\\
Conv7\_1 & Conv3D & (T, 256, 256, 32) & - \\
Conv7\_2 & Conv3D & (T, 256, 256, 32) & - \\
Output & Conv3D & (T, 256, 256, 3) & - \\
\hline
\end{tabular}
\label{tab:unet_architecture}
\end{table}

\begin{figure*}
\setlength{\tabcolsep}{1.5pt}
\centering
\begin{tabular}{cccc}
    \includegraphics[width=0.23\textwidth]{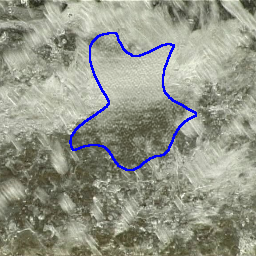}
    & \includegraphics[width=0.23\textwidth]{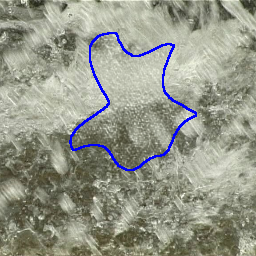}
    & \includegraphics[width=0.23\textwidth]{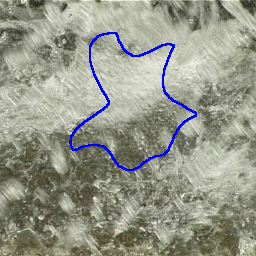}
    & \includegraphics[width=0.23\textwidth]{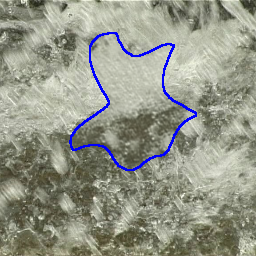} \\
    
    \includegraphics[width=0.23\textwidth]{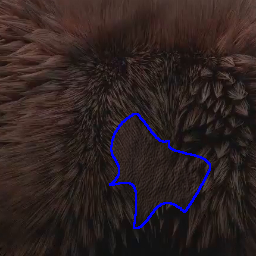}
    & \includegraphics[width=0.23\textwidth]{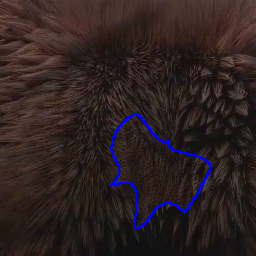}
    & \includegraphics[width=0.23\textwidth]{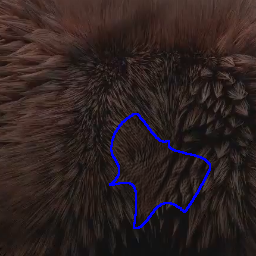}
    & \includegraphics[width=0.23\textwidth]{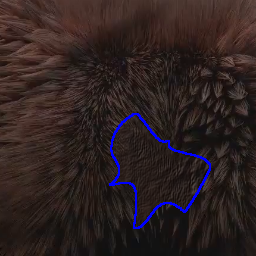} \\
    Interval 1 & Interval 10 & Interval 50 & Interval 100 \\
\end{tabular}
\caption{Inpainting visual results with different interval sizes. Short intervals have a more noisy texture than long intervals. Best seen with zoom-in. \textit{All methods are trained with the same training budget.}}
\label{fig:ablation}
\end{figure*}

\begin{figure*}
    \centering
    \includegraphics[width=0.8\textwidth]{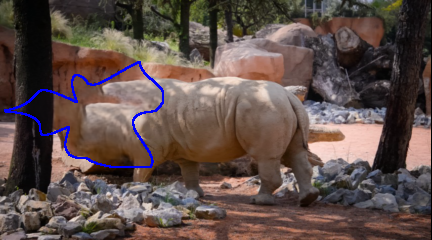}
    \caption{Example of failure case. Semantic inpainting is often impossible due to our learning strategy and limited training data; we assume similarity between inpainted content and visible content. In this case, the rhino's head is not correctly reconstructed.}
    \label{fig:limit_self}
\end{figure*}

\end{document}